\documentclass{aicom2e}
\pdfoutput=1

\usepackage[active]{srcltx}
\usepackage{ amsthm, amssymb,amsfonts}
\usepackage{graphicx}
\usepackage{epsfig}
\usepackage{latexsym}
\usepackage{colortbl}
\usepackage{times}
\usepackage{floatrow}
\usepackage{helvet}
\usepackage{courier}
\usepackage{subfig}
\usepackage{float}
\usepackage{xcolor}
\usepackage{pslatex}
\usepackage{eso-pic}
\usepackage[bottom]{footmisc}
\usepackage{etex}
\usepackage{url}
\usepackage{latexsym}
\usepackage{vntex}
\usepackage[english]{babel}
\usepackage{graphicx}
\usepackage{framed}
\usepackage{multirow}
\usepackage[font=small]{caption}

\usepackage[hidelinks]{hyperref}
\usepackage{enumitem}
\usepackage{fixltx2e}
\usepackage[numbers]{natbib}

\begin{document}
\begin{frontmatter}

\title{A Robust Transformation-Based Learning Approach Using Ripple Down Rules \\ for Part-of-Speech Tagging}

\runningtitle{A Robust Transformation-Based Learning Approach Using Ripple Down Rules for Part-of-Speech Tagging}
 
\author[A]{\fnms{Dat Quoc} \snm{Nguyen}\thanks{Corresponding author. E-mail: dat.nguyen@students.mq.edu.au.}\thanks{The first two authors contributed equally to this work.}},
\author[B]{\fnms{Dai Quoc} \snm{Nguyen}},
\author[D]{\fnms{Dang Duc} \snm{Pham}},
and
\author[C]{\fnms{Son Bao} \snm{Pham}}
\address[A]{Department of Computing, Macquarie University, Australia \\
E-mail: dat.nguyen@students.mq.edu.au}
\address[B]{Department of Computational Linguistics, Saarland University, Germany \\
E-mail: daiquocn@coli.uni-saarland.de}
\address[D]{L3S Research Center, University of Hanover, Germany\\
E-mail: pham@l3s.de}
\address[C]{VNU University of Engineering and Technology, Vietnam National University, Hanoi, Vietnam\\
E-mail: sonpb@vnu.edu.vn}

\maketitle

\begin{abstract}
\ In this paper, we propose a new approach to construct a system of transformation rules for the Part-of-Speech (POS) tagging task. Our  approach is based on
an incremental knowledge acquisition method where rules are stored
in an exception structure and new rules are only added to correct the errors
of existing rules; thus allowing systematic control of the interaction between
the rules. Experimental results  on 13 languages show that our approach is fast in terms of training time and tagging speed. Furthermore, our approach obtains very competitive accuracy in comparison to state-of-the-art POS and morphological taggers. 
\end{abstract}


\end{frontmatter}

\section{Introduction}
\label{Introduction}

POS tagging is one of the most important tasks in Natural Language Processing (NLP) that assigns a tag to each word in a text, which the tag represents the word's lexical category \citep{Gungor2010}. After the text has been tagged or annotated, it can be used in many applications such as  machine translation, information retrieval, information extraction and the like. 

Recently,  statistical and machine learning-based POS tagging methods have become the mainstream ones obtaining state-of-the-art performance. 
However, the learning process of many of them is time-consuming and requires  powerful computers for training. For example, for the task of combined POS and morphological tagging, as reported by \citet{mueller-schmid-schutze:2013:EMNLP},  the taggers SVMTool \citep{Gimenez04} and CRFSuite \citep{CRFsuite:2007}  took 2,454 minutes (about 41 hours) and 9,274 minutes (about 155 hours) respectively to train on a  corpus of 38,727 Czech sentences (652,544 words), using a machine with two Hexa-Core Intel Xeon X5680 CPUs with 3,33 GHz and 144 GB of memory. Therefore, such methods might not be reasonable for individuals having limited computing resources. 
In addition, the tagging speed of many of those systems is relatively slow. For example, as reported by \citet{Moore2014}, the SVMTool, the COMPOST tagger \citep{Spoustova2009} and the UPenn bidirectional tagger \citep{Shen07} respectively achieved the tagging speed of 7700, 2600 and 270 English word tokens per second, using a Linux workstation with Intel Xeon X5550 2.67 GHz processors. So these methods may not be adaptable to the recent large-scale data NLP tasks where the fast tagging speed is necessary.

Turning to the rule-based POS tagging methods, the most well-known method proposed by \citet{Brill95}  automatically learns   transformation-based error-driven rules. 
In the Brill's method, the learning process selects a new rule based on the temporary context which is generated by all the preceding rules; the learning process then applies the new rule to the temporary context to generate a new   context. By repeating this process, a sequentially ordered list of rules is produced, where a rule is allowed to  change the outputs of all the preceding rules, so a word could be relabeled multiple times. Consequently, the Brill's method is slow in terms of training and tagging processes  \citep{Hepple00,Ngai01}.

In this paper, we present a new error-driven approach to automatically
restructure transformation rules in the form of a Single Classification Ripple Down Rules
(SCRDR) tree \citep{ComptonJ90,RichardsD09}. In the SCRDR tree, a new rule can only be added when the tree produces an incorrect output. Therefore, our approach allows the interaction between the rules, where a rule can only change the outputs of some preceding rules in a controlled context.  To sum up, our contributions are:

\begin{itemize}[leftmargin=*]

\item We propose a new transformation-based  error-driven approach for POS and morphological tagging task, using SCRDR.\footnote{Our free open-source implementation namely \textit{RDRPOSTagger}   is available  at \url{http://rdrpostagger.sourceforge.net/}} Our approach obtains  fast performance in both learning and tagging process. For example, in the combined POS and morphological tagging task,  our approach takes an average of 61 minutes (about 1 hour) to complete a 10-fold cross validation-based training  on a corpus of 116K Czech sentences (about 1,957K words), using a computer with Intel Core i5-2400 3.1GHz CPU and 8GB of memory. In addition, in the English POS tagging, our approach achieves a  tagging speed of 279K word tokens per second.  So our approach can be used on computers with limited resources or can be adapted to the  large-scale data NLP tasks.
\medskip

\item We provide empirical experiments on the POS tagging task and the combined POS and morphological tagging task for 13 languages. We compare our approach to two other approaches in terms of running time and accuracy, and show that our robust and language-independent method achieves a very competitive accuracy in comparison to the state-of-the-art results. 

\end{itemize}

The paper is organized as follows: sections \ref{scrdr} and \ref{sec:rdrpos} present the SCRDR methodology and our new approach, respectively. Section \ref{sec:experiment} details the experimental results while Section \ref{sec:relatedwork} outlines the related work. Finally, Section \ref{sec:Conclusion} provides the concluding remarks and  future work.

\begin{figure*}[!ht]
\centering
\includegraphics[width=16cm]{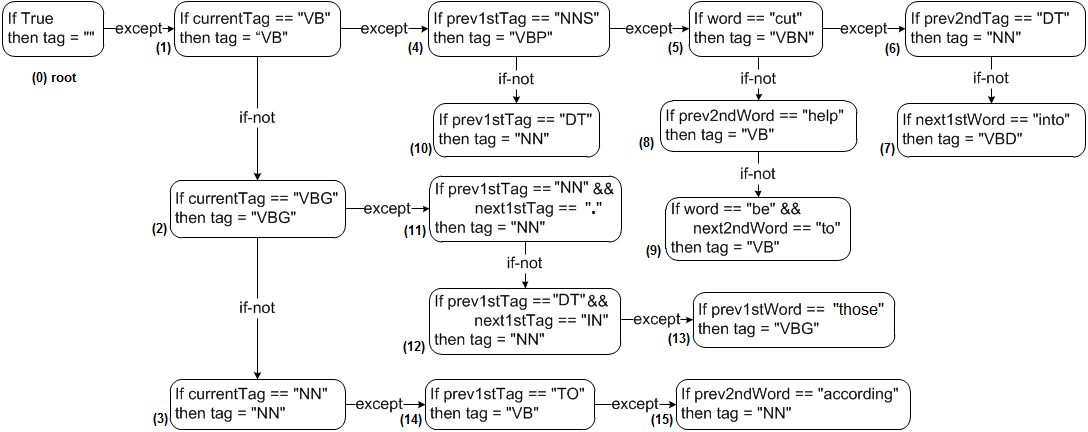} 
\caption{An example of a SCRDR tree for English POS tagging.}
\label{fig:kbscrdr}
\end{figure*}

\section{SCRDR methodology}
\label{scrdr}

A SCRDR tree \citep{ComptonJ90,NguyenNP2015,RichardsD09} is a binary tree with two distinct types of edges.
These edges are typically called {\em except} and {\em if-not} edges.
Associated with each node in the tree is a {\em rule}.
A rule has the form: {\em if
} $\alpha$ {\em then} $\beta$ where $\alpha$ is called the {\em
condition} and $\beta$ is called  the {\em conclusion}.

\begin{table*}[!ht]
\resizebox{16cm}{!}{
\begin{tabular}{l|l}
\hline
Template & Example \\
\hline
\#2: \textit{if}  previous1$^{st}$Word == \textbf{``object.previous1$^{st}$Word''} \textit{then} tag = \textbf{``correctTag''} &  (13)\\
\#3: \textit{if} word == \textbf{``object.word''} \textit{then} tag = \textbf{``correctTag''} &  (5)\\

\#4: \textit{if}  next1$^{st}$Word == \textbf{``object.next1$^{st}$Word''} \textit{then} tag = \textbf{``correctTag''} &  (7)\\
\#10:  \textit{if}  word == \textbf{``object.word''} \&\& next2$^{nd}$Word == \textbf{``object.next2$^{nd}$Word''} \textit{then} tag = \textbf{``correctTag''} &  (9) \\
\#15:  \textit{if}  previous1$^{st}$Tag == \textbf{``object.previous1$^{st}$Tag''} \textit{then} tag = \textbf{``correctTag''} &  (4)\\
\#20: \textit{if}  previous1$^{st}$Tag == \textbf{``object.previous1$^{st}$Tag''} \&\& next1$^{st}$Tag == \textbf{``object.next1$^{st}$Tag''} \textit{then} tag = \textbf{``correctTag''} &  (11) \\
\hline
\end{tabular}
}
\caption{Examples of rule templates corresponding to the  rules (4), (5), (7),  (9), (11) and (13) in Figure \ref{fig:kbscrdr}.}
\label{fig:examrule}
\end{table*} 

Cases in SCRDR
are evaluated by passing a case to the root of the tree.  
At any node in the tree, if the condition of the rule at a node $\eta$ is satisfied by the case (so the node $\eta$ \textit{fires}),
the case is passed on to the {\em except} child node of the node $\eta$ using the {\em except} edge if it exists. Otherwise, the case is passed on to the {\em if-not} child node of the node $\eta$. 
The conclusion of this process is given by the node which \textit{fired} last.

For example,  with the SCRDR tree in  Figure \ref{fig:kbscrdr}, given a case of 5-word window context \textit{``as/IN investors/NNS {anticipate/VB} a/DT recovery/NN''} where \textit{``anticipate/VB''} is the current word and POS tag pair, the case satisfies the conditions of the rules at nodes (0), (1) and (4), then it  is passed on to  node (5), using {\em except} edges. As the case does not satisfy the condition of the rule at node (5), it is passed on to node (8) using the {\em if-not} edge. Also, the case does not satisfy the conditions of the rules at nodes (8) and (9). So we have the evaluation path (0)-(1)-(4)-(5)-(8)-(9) with the last fired node (4). Thus, the POS  tag for \textit{``anticipate''} is concluded as \textit{``VBP''} produced by the rule at node (4).

A new node containing a new exception rule is added to an SCRDR tree when the evaluation process returns an \textit{incorrect} conclusion.  
The new node is attached to the last node in the evaluation path of the given case with the {\em except}
edge if the last node is the {fired} node; otherwise, it is attached with the {\em if-not} edge.
 
 To ensure that a conclusion is always given, the root node (called the {\em default} node) typically contains a trivial condition
which is always satisfied.  
 The rule at the default node, the default rule, is the unique rule which is not an exception rule of any other rule. 

In the SCRDR tree in  Figure \ref{fig:kbscrdr}, rule (1) - the rule at node (1) - is an exception rule of the default rule (0). As node (2) is the {\em if-not} child node of  node (1), rule (2) is also an exception rule of rule (0). Likewise,  rule (3) is  an exception rule of  rule (0). Similarly, both rules (4) and (10) are exception rules of  rule (1) whereas  rules (5), (8) and (9) 
are exception rules of  rule (4), and so on. Therefore, the exception structure of the SCRDR tree extends to four levels: rules (1), (2) and (3) at layer 1; rules (4), (10), (11), (12) and (14) at layer 2; rules (5), (8), (9), (13) and (15) at layer 3; and rules (6) and (7) at layer 4 of the exception structure.

\section{Our approach}
\label{sec:rdrpos}

\begin{figure}[ht]
\centering
\includegraphics[width=7.75cm]{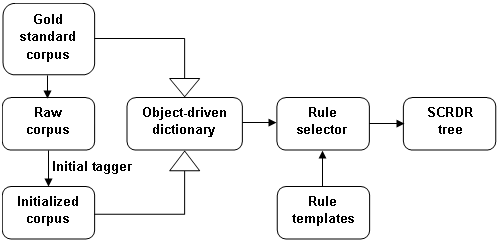} 
\caption{The diagram of our learning process.}
\label{fig:diagram}
\end{figure}

In this section, we present a new error-driven approach to automatically construct a SCRDR tree of transformation rules for POS tagging. The  learning process in our approach is described in Figure \ref{fig:diagram}.

 The \textit{initialized corpus} is generated by using an \textit{initial tagger} to  perform POS tagging on the \textit{raw corpus} which consists of the raw text extracted from the \textit{gold standard training corpus},  excluding POS tags. 

 Our initial tagger uses a lexicon to assign a tag for each word. 
The lexicon is constructed from the gold standard corpus, where each word type is coupled with its most frequent associated tag in the gold standard corpus. 
In addition, the character 2-, 3-, 4- and 5-gram suffixes of word types are also included in the lexicon. 
Each suffix is coupled with the most frequent\footnote{The frequency must be greater than 1, 2, 3 and 4 for the 5-, 4-, 3- and 2-gram suffixes, respectively.} tag associated to the word types containing this suffix. 
Furthermore, the lexicon also contains three default tags corresponding to the tags most frequently assigned to words containing numbers, capitalized words and lowercase words. 
The suffixes  and default tags are only used to label unknown words (i.e. out-of-lexicon words).

To handle  unknown words in English, our initial tagger uses  regular expressions to capture the information about capitalization and word suffixes.\footnote{An example of a regular expression in Python is as follows: \\ \textit{if (re.search(r$'$(.*ness\$) | (.*ment\$) | (.*ship\$) | (\string^\string[Ee\string]x-.*) | (\string^\string[Ss\string]elf-.*)$'$, word) != None): tag = ``NN''}.} For other languages, the initial tagger firstly determines whether the word contains any numeric character to get the default tag for numeric word type. If the word does not contain any numeric character, the initial tagger then extracts the 5-, 4-, 3- and 2-gram suffixes in this order and returns the coupled tag corresponding to the first suffix found in the lexicon. 
If the lexicon does not contain any of the suffixes of the word, the initial tagger determines whether the word is capitalized or in lowercase form  to return the corresponding default tag.

 By comparing the initialized corpus with the {gold standard corpus}, an \textit{object-driven dictionary} of  \textit{Object} and \textit{correctTag} pairs is produced. Each {Object} captures  a 5-word window context of a word and its current initialized tag in the format of (\textit{previous $2^{nd}$ word, previous $2^{nd}$ tag, previous $1^{st}$ word, previous $1^{st}$ tag, word, current tag, next $1^{st}$ word, next $1^{st}$ tag, next  $2^{nd}$ word, next  $2^{nd}$ tag, last-2-characters, last-3-characters, last-4-characters}), extracted from the initialized corpus.\footnote{In the example case from Section 2, the Object corresponding to the 5-word context window is \{\textit{as, IN,  investors, NNS,  anticipate, VB,  a, DT,  recovery, NN, te, ate, pate}\}.}  
The {correctTag} is the corresponding ``true'' tag of the word in the gold standard corpus. 

\begin{table}[ht]
\setlength{\tabcolsep}{0.25em}
\resizebox{7.5cm}{!}{
\begin{tabular}{l|l }
\hline 
words & w\textsubscript{-2}, w\textsubscript{-1}, w\textsubscript{0}, w\textsubscript{+1}, w\textsubscript{+2}  \\ 
\hline 
word bigrams & (w\textsubscript{-2}, w\textsubscript{0}), (w\textsubscript{-1}, w\textsubscript{0}), (w\textsubscript{-1}, w\textsubscript{+1}),
(w\textsubscript{0}, w\textsubscript{+1}) \\
&  (w\textsubscript{0}, w\textsubscript{+2}) \\
\hline
word trigrams &   (w\textsubscript{-2}, w\textsubscript{-1}, w\textsubscript{0}), (w\textsubscript{-1}, w\textsubscript{0}, w\textsubscript{+1}), (w\textsubscript{0}, w\textsubscript{+1}, w\textsubscript{+2}) \\
\hline
POS tags & p\textsubscript{-2}, p\textsubscript{-1},  p\textsubscript{0}, p\textsubscript{+1}, p\textsubscript{+2} \\
\hline
POS bigrams & (p\textsubscript{-2}, p\textsubscript{-1}), (p\textsubscript{-1}, p\textsubscript{+1}), (p\textsubscript{+1}, p\textsubscript{+2}) \\
\hline
Combined & (p\textsubscript{-1}, w\textsubscript{0}), (w\textsubscript{0}, p\textsubscript{+1}),  (p\textsubscript{-1}, w\textsubscript{0}, p\textsubscript{+1}),   (p\textsubscript{-2}, p\textsubscript{-1}, w\textsubscript{0}) \\&  (w\textsubscript{0}, p\textsubscript{+1}, p\textsubscript{+2})  \\
\hline
suffixes & c\textsubscript{n-1}c\textsubscript{n}, c\textsubscript{n-2}c\textsubscript{n-1}c\textsubscript{n}, c\textsubscript{n-3}c\textsubscript{n-2}c\textsubscript{n-1}c\textsubscript{n} \\
\hline
\end{tabular} 
}
\caption{Short descriptions of rule templates. ``w'' refers to word token and ``p'' refers to POS label while -2, -1, 0, 1, 2 refer to indices, for instance, p\textsubscript{0} indicates the current initialized tag. c\textsubscript{n-1}c\textsubscript{n}, c\textsubscript{n-2}c\textsubscript{n-1}c\textsubscript{n}, c\textsubscript{n-3}c\textsubscript{n-2}c\textsubscript{n-1}c\textsubscript{n} correspond to the character 2-, 3- and 4-gram suffixes of w\textsubscript{0}. So the templates \#2, \#3, \#4, \#10, \#15 and \#20 in Table \ref{fig:examrule} are associated to  w\textsubscript{-1}, w\textsubscript{0}, w\textsubscript{+1},  (w\textsubscript{0}, w\textsubscript{+2}), p\textsubscript{-1} and (p\textsubscript{-1}, p\textsubscript{+1}), respectively.} 
\label{tab:templates}
\end{table}

 The \textit{rule selector} is responsible for selecting the most suitable rules to build the \textit{SCRDR tree}. To generate concrete rules, the {rule selector} uses \textit{rule templates}. The examples of our rule templates are presented in Table \ref{fig:examrule}, where the elements in \textbf{bold} will be replaced by specific values from the Object and correctTag pairs in the object-driven dictionary.  Short descriptions of the rule templates are shown in Table \ref{tab:templates}.

 The {SCRDR rule tree} is  initialized with the default rule  \emph{\textbf{if} True \textbf{then} tag = ``''}  as shown in Figure \ref{fig:kbscrdr}.\footnote{The default rule returns an incorrect conclusion of empty POS tag for every Object.} 
Then the system creates a rule of the form \emph{\textbf{if} currentTag == ``\textbf{Label}'' \textbf{then} tag = ``\textbf{Label}''} for each POS tag in the list of all tags extracted from the initialized corpus. These rules are added to the SCRDR tree as exception rules of the default rule to create the first layer exception structure, as for instance the rules (1), (2) and (3) in Figure \ref{fig:kbscrdr}.

\subsection{Learning process} 

The process to construct new exception rules to higher layers of the exception structure in the SCRDR tree is as follows:

\begin{itemize}[leftmargin=*]

\item At each node $\eta$ in the SCRDR tree, let $\Theta_{\eta}$ be the set of Object and correctTag pairs from the object-driven dictionary such that the node $\eta$ is the last fired node for every Object in $\Theta_{\eta}$ and the  node $\eta$ returns an incorrect POS tag (i.e. the POS tag concluded by the node $\eta$ for each Object in $\Theta_{\eta}$ is not the corresponding {correctTag}). A new exception rule must be added to the next level of the SCRDR tree to correct the errors given by the node $\eta$.

\item The new exception rule  is selected from all concrete rules generated  for all Objects in $\Theta_{\eta}$. The selected rule must satisfy the following constraints: 
(i) If node $\eta$ is at level-$k$ exception structure in the SCRDR tree such that $k > 1$ then the rule's condition must not be satisfied by the Objects for which node $\eta$ has already returned a correct POS tag.
(ii) Let $A$ and $B$ be the number of Objects in $\Theta_{\eta}$ that satisfy the rule's condition, and the rule's conclusion returns the correct and incorrect POS tag, respectively. Then the rule with the highest score value $S = A - B$ will be chosen. 
  (iii) The score $S$ of the chosen rule must be higher than a given threshold. We  apply two threshold parameters: the first threshold is to find exception rules  at the layer-2 exception structure, such as rules (4), (10) and (11) in Figure \ref{fig:kbscrdr}, while the second threshold is to find rules for higher exception layers.

\item If the learning process is unable to select a new exception rule, the learning process is repeated at   node $\eta_\rho$ for which the rule at the node $\eta$ is an exception rule of the rule at  the node $\eta_\rho$. Otherwise, the learning process is repeated at the new selected exception rule.

\end{itemize}

\textbf{Illustration:}  To {illustrate} how  new exception rules are added to build a SCRDR tree in Figure \ref{fig:kbscrdr}, we start with node (1) associated to rule (1) \textit{\textbf{if} currentTag == ``VB''  \textbf{then} tag = ``VB''} at the layer-1 exception structure. The learning process chooses the rule  \textit{\textbf{if} prev1$^{st}$Tag == "NNS" \textbf{then} tag = "VBP"} as an exception rule for rule (1). Thus, node (4) associated with  rule (4) \textit{\textbf{if} prev1$^{st}$Tag == "NNS" \textbf{then} tag = "VBP"} is added as an {\em except} child node of  node (1). The learning process is then repeated at  node (4). Similarly, nodes (5) and (6) are added to the tree as shown in Figure  \ref{fig:kbscrdr}.

 The learning process now is repeated at  node (6). At  node (6),  the learning process cannot find a suitable rule that satisfies the three constraints described above. So the learning process is repeated at node (5) because  rule (6) is an exception rule of  rule (5). 
At  node (5), 
 the learning process selects a new rule (7) \textit{\textbf{if} next1$^{st}$Word == "into" \textbf{then} tag = "VBD"} to be another exception rule of  rule (5). Consequently, a new node (7) containing rule (7) is added to the  tree as an {\em if-not} child node of  node (6).  
At node (7), the learning process cannot find a new rule to be an exception rule of  rule (7). Therefore, the learning process is again repeated at node (5). 

\medskip

\textit{This process of adding new exception rules is repeated until no rule satisfying the three constraints can be found.}

\subsection{Tagging process} 
The tagging process firstly tags unlabeled text by using the initial tagger.  
Next, for each initially tagged word the corresponding Object will be created by sliding a 5-word context window over the text from left to right. 
Finally, each word will be tagged by passing its Object through the learned SCRDR tree, as illustrated in the example in Section \ref{scrdr}. 
 If the default node is the last fired node satisfying the Object, the final tag returned is the tag produced by the initial tagger.


\begin{table*}[!ht]

{\small
\begin{tabular}{ll|l|l|l|l|l}
\hline 
\bf Language & \bf Source & \bf \#sen & \bf \#words & \bf \#P & \bf \#PM  &  \bf OOV\\ 
\hline 
\hline
Bulgarian 
& BulTreeBank-Morph \citep{simov2004}  & 20,558 & 321,538 & ---& 564 & 10.07 \\
\hline
Czech & PDT Treebank  2.5 \citep{Bejcek12} & 115,844 & 1,957,246 & --- & 1,570 & 6.09 \\
\hline
Dutch 	& Lassy Small Corpus \citep{Gertjan:2013} & 65,200 & 1,096,177 & --- & 933 & 7.21 \\
\hline
French & French Treebank \citep{Anne:2003} & 21,562 & 587,687 & 17 & 306 & 5.19 \\
\hline 
German & TIGER Corpus \citep{Brants04} & 50,474 & 888,236 & 54 & 795 & 7.74 \\
\hline
Hindi & Hindi Treebank \citep{Martha09} & 26,547 & 588,995 & 39 & --- & --- \\
\hline
Italian &  ISDT Treebank \citep{Bosco2013} & 10,206 & 190,310 & 70 & --- &  11.57 \\
\hline
Portuguese & Tycho Brahe Corpus \citep{galves10}  & 68,859 & 1,482,872 & --- & 344 & 4.39\\
\hline
Spanish & IULA LSP Treebank \citep{MARIMON12} & 42,099 & 589,542 & --- & 241 & 4.94 \\
\hline
Swedish & Stockholm—Umeå Corpus 3.0 \citep{suc30}  & 74,245 & 1,166,593 & ---  
& 153 & 8.76\\
\hline
Thai & ORCHID Corpus \citep{Charoenporn} & 23,225 & 344,038 & 47 & --- & 5.75 \\
\hline
Vietnamese & (VTB) Vietnamese Treebank \citep{nguyen-EtAl:2009:LAW-III} & 10,293 & 220,574 & 22 & --- & 3.41\\
 & (VLSP) VLSP Evaluation Campaign 2013 & 28,232 & 631,783 & 31 & --- & 2.06\\
\hline
\hline
\end{tabular} 
}
\caption{The experimental datasets. \textbf{\#sen}: the number of sentences. \textbf{\#words}: the  number of words. \textbf{\#P}: the number of POS tags. \textbf{\#PM}: the number of combined POS and morphological (POS+MORPH) tags. \textbf{OOV} (Out-of-Vocabulary): the average percentage of unknown word tokens in each test fold. For Hindi, OOV rate is 0.0\% on 9 test folds while it is 3.8\% on the remaining test fold.}
\label{tab:sources}
\end{table*}

\section{Empirical study}
\label{sec:experiment}

This section presents the experiments validating our proposed approach in 13 languages. We also compare our approach with the TnT\footnote{\url{www.coli.uni-saarland.de/~thorsten/tnt/}} approach \citep{Brants00} and the MarMoT\footnote{\url{http://cistern.cis.lmu.de/marmot/}}  approach  proposed by  \citet{mueller-schmid-schutze:2013:EMNLP}. The  TnT tagger is considered as one of the fastest POS taggers in
literature (both in terms of training and tagging), obtaining competitive  tagging accuracy on diverse languages \citep{Gungor2010}. The MarMoT tagger is a morphological tagger obtaining state-of-the-art tagging accuracy on various languages such as Arabic, Czech, English, German, Hungarian and Spanish.

We run all experiments on a computer of Intel Core i5-2400 3.1GHz CPU and 8GB of memory. Experiments on English use the Penn WSJ Treebank \citep{Marcus93}  sections 0-18 (38,219 sentences - 912,344 words) for training,   sections 19-21 (5,527 sentences - 131,768 words) for validation, and the  sections 22-24  (5,462 sentences - 129,654 words) for testing. The proportion of unknown words in the test set is 2.81\% (3,649 unknown words). 
We also conduct experiments on 12 other languages. The experimental datasets for those languages are described in Table \ref{tab:sources}.

Apart from English, it is difficult to compare the results of previously published works because each of them have used different experimental setups and data splits. Thus, it is difficult to create the same evaluation settings used in the previous works. 
So we perform 10-fold cross validation\footnote{For each dataset, we split the dataset into 10 contiguous parts (i.e. 10 contiguous folds).  The evaluation procedure is repeated 10 times. 
Each part is used as the test set and 9 remaining parts are merged as the training set.
\textit{All accuracy results are reported as the average results over the test folds}.} for all languages other than English, except for Vietnamese where we use 5-fold cross validation.

{\textbf{Our approach}: }  In training phase, all words appearing only once time in the training set are initially treated as unknown words and tagged as described in Section \ref{sec:rdrpos}. This strategy produces tagging
models containing transformation rules learned 
on error contexts of unknown words. The threshold parameters were tuned on the English validation set. The best value pair (3, 2) was then used in all experiments for all languages.

\textbf{TnT} \& \textbf{MarMoT}: We used default parameters for training  TnT and MarMoT. 


 

\begin{table*}[ht]

\setlength{\tabcolsep}{0.5em}
\resizebox{16.0cm}{!}{
{\small
\begin{tabular}{l|lll|lll|ll|lll|ll}
\hline
& \multicolumn{8}{c|}{\bf RDRPOSTagger}  & \multicolumn{5}{c}{\bf TnT} \\ 
\cline{2-14}
\bf Language & \multicolumn{3}{c|}{Initial accuracy} & \multicolumn{3}{c|}{Tagging accuracy} & \multicolumn{2}{c|}{Speed} & \multicolumn{3}{c|}{Tagging accuracy} & \multicolumn{2}{c}{Speed}   \\
\cline{2-14}
& Kno. & Unk. & All.  & Kno. & Unk. & All. & TT & TS & Kno. & Unk. & All. & TT & TS \\
\hline 
\hline
Bulgarian$^*$ &  95.13 & 49.50 & 90.53 & 96.59 & 66.06 & 93.50 & 2 & 157K & 96.55 &  70.10 &  \textbf{93.86}$^+$ & 1 & 313K \\
\hline
Czech$^*$ & 84.05 & 52.60 & 82.13 & 93.01 & 64.86 & 91.29 & 61 & 56K & 92.95 &  67.83 & \textbf{91.42}$^+$ & 1 & 164K \\
\hline
Dutch$^*$ & 88.91 & 54.30 & 86.34  & 93.88 & 60.15 & 91.39 & 44  & 103K & 93.32 &  69.07 &  \textbf{91.53} & 1 & 125K \\
\hline
English & 93.94 & 78.84 & 93.51  & 96.91 & 83.89 & {\textbf{96.54}$^+$} & 18 & 279K & 96.77 &  86.02 &  96.46 & 1 & 720K \\
\hline
French & 95.99 & 77.18 & 94.99 & 98.07 & 81.57 & \textbf{97.19} & 16 & 237K & 97.52 &  87.43 & {96.99} & 1 & 722K \\
\hline
French$^*$ & 89.97 & 54.36 & 88.12 & 95.09 & 63.74 & 93.47 & 9  & 240K & 95.13 & 70.67 &  \textbf{93.88}$^+$ & 1 & 349K \\
\hline
German & 94.76 & 73.21 & 93.08 & 97.74 & 78.87 & 96.28  & 28 & 212K & 97.70 &  89.38 & \textbf{97.05}$^+$ & 1 & 509K \\
\hline
German$^*$ &  71.68  & 30.92  & 68.52 & 87.70  & 51.84  & 84.92 & 22  & 111K &  86.98 &  61.22 &  \textbf{84.97} & 1 & 98K \\
\hline
Hindi & \_\_ & \_\_ & 89.63 & \_\_ & \_\_ & \textbf{95.77}$^+$ & 21 & 210K & \_\_ &  \_\_ &  94.80 & 1& 735K \\
\hline
Italian & 92.63 & 67.33 & 89.59  & 95.93 & 71.79 & 93.04 & 3 & 276K & 96.38 &   86.16 & \textbf{95.16}$^+$ & 1  & 446K \\
\hline
Portuguese$^*$ &  92.85 & 61.19 & 91.43  & 96.07 & 64.38 & 94.66 & 42 & 172K & 96.01 &  78.81 & \textbf{95.24}$^+$ & 1 & 280K\\
\hline
Spanish$^*$ & 97.94 & 75.63 & 96.92 & 98.85 & 79.50  & 97.95  & 4  & 283K &  98.96 &  84.16 &  \textbf{98.18} & 1 & 605K\\
\hline
Swedish$^*$ &  90.85  & 71.60 & 89.19  & 96.41 & 76.04  & 94.64 & 41  & 152K & 96.33 &  85.64 &  \textbf{95.39}$^+$ & 1  & 326K\\ 
\hline
Thai &  92.17  & 75.91 & 91.23  & 94.98  & 80.68 & \textbf{94.15}$^+$ & 6 & 315K & 94.32 & 80.93 &  93.54 & 1 & 490K \\
\hline
Vn (VTB) & 92.17 & 55.21 & 90.90  & 94.10 & 56.38 & \textbf{92.80}$^+$ & 5 & 269K & 92.90 & 59.35 & 91.75 & 1 & 723K \\
\ \ \ \ \ \ (VLSP) & 91.88 &  64.36 &  91.31 & 94.12 & 65.38 & \textbf{93.53}$^+$ & 23 & 145K &  92.65 & 68.07 & 92.15 & 1 &  701K \\
\hline
\hline
\end{tabular} 
}
}
\caption{The accuracy results (\%) of our approach using the lexicon-based initial tagger (for short, RDRPOSTagger) and TnT. Languages marked with * indicate the tagging accuracy on combined POS+MORPH tags. \textbf{Vn} $\rightarrow$ Vietnamese.
\textbf{Kno.}: the  known word tagging accuracy. \textbf{Unk.}: the unknown word tagging accuracy. \textbf{All.}: the overall accuracy result. \textbf{TT}:  training time (minutes). \textbf{TS}:  tagging speed (number of word tokens per second).  Higher results are highlighted in \textbf{bold}. Results marked \textbf{$^+ $} refer to a significant test with p-value < 0.05, using the two sample Wilcoxon test; due to a non-cross validation evaluation, we used accuracies over POS labels to perform significance test for English. }
\label{tab:acc1}
\end{table*}

\begin{table*}[ht]
{\small
\begin{tabular}{l|lll|lll|ll}
\hline
& \multicolumn{3}{c|}{\bf RDRPOSTagger\textsubscript{+TnT}}  & \multicolumn{5}{c}{\bf MarMoT } \\ 
\cline{2-9}
\bf Language &   \multicolumn{3}{c|}{Tagging accuracy}   & \multicolumn{3}{c|}{Tagging accuracy} & \multicolumn{2}{c}{Speed}  \\
\cline{2-9}
 & Kno. & Unk. & All. & Kno. & Unk. & All. & TT & TS \\
\hline 
\hline
Bulgarian$^*$  & 96.82 & 70.27 & 94.12 &   96.92 & 76.72 & \textbf{94.86}$^+$ & 9 & 4K \\
\hline
Czech$^*$   & 93.24 & 67.92 &  91.70  & 94.74 &  75.84 & \textbf{93.59}$^+$ & 130 & 2K\\
\hline
Dutch$^*$  & 94.00&  69.20 &  92.17  & 94.74 &  73.39 &  \textbf{93.17}$^+$  & 44 & 3K \\
\hline
English   & 97.17 & 86.19 & 96.86  & 97.47 &  89.39 &  \textbf{97.24} & 5 & 16K \\
\hline
French   &  98.27 & 87.55 &  97.70   & 98.33 &  91.15 &  \textbf{97.93} & 2 & 12K \\
\hline
French$^*$   & 95.42 &  70.93 &  94.16   & 95.55 &  77.66 &  \textbf{94.62}$^+$ & 9 & 6K \\
\hline
German   & 98.13 &  89.43 & 97.46  & 98.30 & 92.54 &  \textbf{97.85}$^+$ & 5 & 9K \\
\hline
German$^*$   & 87.65 &  62.05 & 85.66  & 90.61 &  69.13 &  \textbf{88.94}$^+$ & 32 & 3K \\
\hline
Hindi & \_\_ & \_\_ &  96.21  & \_\_ & \_\_ & \textbf{96.61}$^+$ & 3 & 16K \\
\hline
Italian   & 96.75 &  86.18 &  95.49  &  96.90 & 89.21 & \textbf{95.98}$^+$ & 2 & 6K \\
\hline
Portuguese$^*$  & 96.30 & 78.81 &  95.53  & 96.53 & 81.49 &  \textbf{95.86}$^+$ &  23 & 6K \\
\hline
Spanish$^*$  &  99.05 & 84.13 & 98.26  & 99.08 &  86.86 &  \textbf{98.45}$^+$ & 8 & 8K \\
\hline
Swedish$^*$   & 96.79 &  85.68 &  95.81  &  97.15 &   86.63 &  \textbf{96.22}$^+$ & 11 & 7K \\ 
\hline
Thai   &  95.03 & 81.10 & 94.21  &  95.42 & 86.99 &  \textbf{94.94}$^+$ & 2 & 12K \\
\hline
Vn (VTB)   & 94.15 & 59.39 &  92.95  & 94.37 & 69.89 &  \textbf{93.53}$^+$ & 1 & 16K \\
\ \ \ \ \ \ (VLSP)   & 94.16 &  68.14 &  93.63  & 94.52 &  75.36 &  \textbf{94.13}$^+$ & 3 & 21K \\
\hline
\hline
\end{tabular} 
}
\caption{The accuracy results (\%) of our approach using TnT as the initial tagger (for short, RDRPOSTagger\textsubscript{+TnT}) and MarMoT. }
\label{tab:acc2}
\end{table*}

\subsection{Accuracy Results}
\label{sec:accresults}

We present the tagging accuracy of our  approach with the lexicon-based initial tagger (for short, RDRPOSTagger) and TnT  in Table  \ref{tab:acc1}. 
As can be seen from Table \ref{tab:acc1}, our RDRPOSTagger does better than TnT on isolating languages such as Hindi, Thai and Vietnamese. For the combined POS and morphological (POS+MORPH) tagging task on  morphologically rich languages such as Bulgarian, Czech, Dutch, French, German, Portuguese, Spanish and Swedish, RDRPOSTagger and TnT generally obtain similar results on known words. However, RDRPOSTagger performs worse on unknown words. 
This can be because RDRPOSTagger uses a simple lexicon-based method for tagging unknown words, while TnT uses a more complex suffix analysis to handle unknown words. Therefore, TnT performs better than RDRPOSTagger on morphologically rich languages.

These initial accuracy results could be improved by following any of the previous studies that use external lexicon resources or existing morphological analyzers. In this research work, we simply  employ TnT as the initial tagger in our approach.  We report the accuracy results of our approach using TnT as the initial tagger (for short, RDRPOSTagger\textsubscript{+TnT}) and MarMoT in Table \ref{tab:acc2}. To sum up, RDRPOSTagger\textsubscript{+TnT} obtains  competitive  results in comparison to the state-of-the-art MarMoT tagger, across the 13 experimental languages. In particular, excluding Czech and German where MarMoT embeds existing morphological analyzers, RDRPOSTagger\textsubscript{+TnT} obtains accuracy results which mostly are about 0.5\%  lower than MarMoT's.

\subsubsection{English}

RDRPOSTagger produces a SCRDR tree model of 2,549 rules in a 5-level exception structure and achieves an accuracy of 96.54\% against 96.46\% accounted for  TnT, as presented in Table \ref{tab:acc1}. 
Table \ref{tab:accEnglish} presents the accuracy results obtained up to each exception level of the tree. 

\begin{table}[ht]
\begin{tabular}{lll}
\hline Level & 	Number of rules	& Accuracy\\
 \hline
 $<=$ 1	& 47	& 93.51 \% \\ 
  $<=$ 2	& 1,522	& 96.36 \% \\ 
  $<=$ 3 &	2,503	& 96.53	\% \\ 
  $<=$ 4	& 2,547	& 96.54	\% \\ 
  $<=$ 5	& 2,549	& 96.54	\% \\ 
\hline 
\end{tabular} 
\caption{Results  due to levels of exception structures.}
\label{tab:accEnglish}
\end{table}

As shown in \cite{nguyen2011ripple}, using the same evaluation scheme for English,  the Brill's rule-based tagger V1.14 \cite{Brill95} gained a similar accuracy result at 96.53\%.\footnote{The Brill's tagger uses an initial tagger with an accuracy of 93.58\% on the test set. Using this initial tagger, our approach gains a higher accuracy
of 96.57\%.} Using TnT as the initial tagger, RDRPOSTagger\textsubscript{+TnT} achieves an accuracy of 96.86\% which is comparable to the state-of-the-art result at 97.24\% obtained by MarMoT.

\subsubsection{Bulgarian}

In Bulgarian, RDRPOSTagger\textsubscript{+TnT} obtains an accuracy of 94.12\% which is 0.74\% lower than the accuracy of MarMoT at 94.86\%.

This is better than the results reported on the BulTreeBank webpage\footnote{http://www.bultreebank.org/taggers/taggers.html} on POS+MORPH tagging task, where  TnT,  SVMTool \citep{Gimenez04} and the memory-based tagger in the Acopost package\footnote{http://acopost.sourceforge.net/} \citep{Schroder2002} obtained accuracies of 92.53\%, 92.22\% and 89.91\%, respectively. 
 Our result is also better than the accuracy of 90.34\% reported by \citet{Georgiev2009}, obtained with the Maximum Entropy-base POS tagger from the OpenNLP toolkit.\footnote{http://opennlp.sourceforge.net}
 Recently,
\citet{Georgiev2012}\footnote{\citet{Georgiev2012} split the BulTreeBank corpus into training set of 16,532 sentences, development set of 2,007 sentences and test set of 2,017 sentences.} reached the state-of-the-art  accuracy result of 97.98\% for POS+MORPH tagging, however, without external resources  the accuracy was 95.72\%. 

\subsubsection{Czech} 
 
\citet{mueller-schmid-schutze:2013:EMNLP} presented the results of five POS taggers SVMTool, CRFSuite  \citep{CRFsuite:2007}, RFTagger \citep{Schmid:2008},  Morfette \citep{CHRUPALA08} and MarMoT for Czech POS+MORPH tagging. All models were trained using a training set of 38,727 sentences (652,544 tokens) and evaluated on a test set of 4,213 sentences (70,348 tokens), extracted from the Prague Dependency Treebank 2.0. The accuracy results  are 89.62\%, 90.97\%, 90.43\%, 90.01\% and 92.99\% accounted for  SVMTool, CRFSuite, RFTagger, Morfette and MarMoT, respectively.

Since we could not access the Czech datasets used in the experiments above, we employ the Prague Dependency Treebank 2.5 \citep{Bejcek12} containing about 116K sentences. The accuracies of  RDRPOSTagger (91.29\%) and RDRPOSTagger\textsubscript{+TnT} (91.70\%) compare  favorably to the result of MarMot (93.50\%).

\subsubsection{Dutch} 

The TADPOLE tagger \cite{Bosch2007} was reached an accuracy of 96.5\% when trained on a manually POS-annotated corpus containing 11 million Dutch words and 316 tags. Due to the limited access we could not use this corpus in our experiments and thus we can not compare our results with the TADPOLE tagger. Instead, we use the Lassy Small Corpus \citep{Gertjan:2013} containing about 1.1 million words.
 RDRPOSTagger\textsubscript{+TnT} achieves a promising  accuracy at 92.17\% which is 1\% absolute lower than the accuracy of MarMoT (93.17\%).

 \subsubsection{French} 

Current state-of-the-art methods for French POS tagging have reached accuracies up to 97.75\% \citep{Seddah2010,Denis:2012}, using the French Treebank \citep{Anne:2003} with  9,881 sentences for training and 1,235 sentences for test. However, these methods  employed Lefff  \citep{Sagot2006} which is an external large-scale morphological lexicon. Without using the lexicon, \citet{Denis:2012} reported an accuracy performance at 97.0\%.
 
 We trained our systems on 21,562 annotated French Treebank sentences and gained a POS tagging accuracy of 97.70\% using RDRPOSTagger\textsubscript{+TnT} model, which is comparable to the  accuracy at 97.93\% of MarMoT.  Regarding to  POS+MORPH tagging, as far as we know  this is the first experiment for French, where RDRPOSTagger\textsubscript{+TnT}  obtains an accuracy  of 94.16\% against 94.62\% obtained by MarMoT.

 \subsubsection{German} 

Using the 10-fold cross validation  evaluation scheme on the TIGER corpus \citep{Brants04} of  50,474 German sentences, \citet{Giesbrecht2007} presented the results of TreeTagger \citep{Schmid1994a}, TnT, SVMTool, Stanford tagger \citep{Toutanova2003}  and Apache UIMA Tagger\footnote{https://uima.apache.org/sandbox.html\#tagger.annotator} obtaining the POS tagging accuracies at 96.89\%, 96.92\%, 97.12\%, 97.63\% and 96.04\%, respectively. In the same evaluation setting, RDRPOSTagger\textsubscript{+TnT} gains an accuracy result of 97.46\% while MarMoT gains a higher accuracy at 97.85\%.      

Turning to POS+MORPH tagging, \citet{mueller-schmid-schutze:2013:EMNLP} also performed  
 experiments on the TIGER corpus,  using   40,474 sentences for training and  5,000 sentences for test.  
They presented accuracy performances of 83.42\%, 85.68\%, 84.28\%, 83.48\% and 88.58\% obtained with the taggers SVMTool, CRFSuite, RFTagger, Morfette and MarMoT, respectively. 
In our evaluation scheme, RDRPOSTagger and RDRPOSTagger\textsubscript{+TnT} correspondingly achieve favorable accuracy results at 84.92\% and 85.66\% in comparison to an accuracy at 88.94\% of MarMoT. 

 \subsubsection{Hindi} 
 
On the Hindi Treebank \citep{Martha09}, RDRPOSTagger\textsubscript{+TnT}  reaches a competitive accuracy result of  96.21\% against the accuracy of MarMoT at 96.61\%. Being one of the largest languages in the world, there are many previous works on POS tagging for Hindi. However, most of them have used small manually labeled datasets that are not publicly available and that are smaller than the Hindi Treebank used in this paper.

\citet{Joshi2013} achieved an accuracy of 92.13\% using a Hidden Markov Model-based approach, trained on a dataset of 358K words and tested on 12K words. 
Using another training set of 150K words and test set of 40K words, \citet{Agarwal2011} compared machine learning-based approaches and presented the  POS tagging accuracy at 93.70\%.

In the 2007 Shallow Parsing Contest for South Asian Languages \citep{Bharathi2007}, the POS tagging track provided a small training set of 21,470 words and a test set of 4,924 words. The highest accuracy in the contest was 78.66\% obtained by \citet{Pvs2007}. In the same 4-fold cross validation evaluation scheme using a dataset of 15,562 words, \citet{Singh2006} obtained an  accuracy of 93.45\% whilst \citet{Dalal07} achieved a result at 94.38\%.

  \subsubsection{Italian}
  
  In the EVALITA 2009 workshop on Evaluation of NLP and Speech Tools for Italian\footnote{http://www.evalita.it/2009}, the POS tagging track \citep{Attardi2009}  provided a training set of 3,719 sentences (108,874 word forms) with 37 POS tags. The teams participating in the closed task where using external resources was not allowed  achieved various tagging accuracies  on a test set of 147 sentences (5,066 word forms), ranging from 93.21\% to 96.91\%.

 Our experiment on Italian POS tagging employs the ISDT Treebank \citep{Bosco2013} of 10,206 sentences (190,310 word forms) with 70 POS tags. RDRPOSTagger\textsubscript{+TnT} obtains a competitive accuracy performance at  95.49\% against 95.98\% computed for MarMoT.

 \subsubsection{Portuguese}

The previous works \citep{Janeiro2008,Kepler:2006} on  POS+MORPH tagging for Portuguese used an early version of the  Tycho Brahe corpus \citep{galves10} containing about 1,036K words. The corpus was split into a training set of 776K words and a test set of 260K words. Based on this setting, \citet{Kepler:2006} achieved an accuracy of 95.51\% while \citet{Janeiro2008} reached a state-of-the-art accuracy result at 96.64\%. 

The Tycho Brahe corpus  in our experiment consists of about 1,639K words. RDRPOSTagger\textsubscript{+TnT} reaches an accuracy at  95.53\% while MarMoT obtains higher result at 95.86\% on 10-fold cross validation.
 
\subsubsection{Spanish} 
 
In addition to Czech and German, \citet{mueller-schmid-schutze:2013:EMNLP}  evaluated the five taggers of  SVMTool, CRFSuite, RFTagger,  Morfette and MarMoT for Spanish POS+MORPH tagging, using a training set of 14,329 sentences (427,442 tokens) and a test set of 1,725 sentences (50,630 tokens) with 303 POS+MORPH tags. The  accuracy results of the five taggers ranged from 97.35\% to 97.93\%, in which MarMoT obtained the highest result.

As we could not access the training and test sets used in \citet{mueller-schmid-schutze:2013:EMNLP}'s experiment, we use the  IULA Spanish LSP Treebank  \citep{MARIMON12} of 42K sentences with 241 tags. RDRPOSTagger and RDRPOSTagger\textsubscript{+TnT} achieve accuracies of 97.95\% and 98.26\%, respectively, while MarMoT obtains a higher result at 98.45\%.

\textbf{NOTE} that here we can make an indirect comparison between our RDRPOSTagger and the SVMTool, CRFSuite, RFTagger and  Morfette taggers via MarMoT. We conclude that the results of  RDRPOSTagger  would likely be similar to the results of SVMTool, CRFSuite, RFTagger and  Morfette on Spanish as well as on Czech and German.

\subsubsection{Swedish} 

On the same SUC corpus 3.0 \citep{suc30}   consisting of 500 text files with about 74K sentences that we also use, \citet{Ostling13}  evaluated the Swedish POS tagger Stagger using 10-fold cross validation but the folds were split at the file level and not on sentence level as we do. Stagger attained an accuracy of 96.06\%.

In our experiment, RDRPOSTagger\textsubscript{+TnT} obtains an accuracy result of 95.81\% in comparison to the accuracy at 96.22\% of MarMoT. 

\subsubsection{Thai} 

On the  Thai POS Tagged  corpus ORCHID \citep{Charoenporn}  of 23,225 sentences, RDRPOSTagger\textsubscript{+TnT} achieves an accuracy of 94.22\% which is 0.72\% absolute lower than the accuracy result of MarMoT (94.94\%). 

It is difficult to compare our results to the previous work on Thai POS tagging. For example, the previous works \citep{Ma2000,Murata2002} performed their experiments on an unavailable corpus of 10,452 sentences. The ORCHID corpus was also used in a POS tagging experiment presented by \citet{Kruengkrai2006}, however, 
the obtained accuracy of 79.342\% was dependent on the performance of automatic word segmentation. On another corpus of 100K words,
  \citet{Pailai13} reached an accuracy of 93.64\% using 10-fold cross validation. 

 \subsubsection{Vietnamese}
 
  We participated in the first evaluation campaign on Vietnamese language processing\footnote{http://uet.vnu.edu.vn/rivf2013/campaign.html} (VLSP).  The campaign's POS tagging track provided a  training set of 28,232 POS-annotated sentences   and an unlabeled test set of 2,130 sentences. 
  RDRPOSTagger achieved the 1$^{st}$ place in the POS tagging track.
 
In this paper, we also carry out POS tagging experiments using  5-fold cross validation evaluation scheme on the VLSP set of 28,232 sentences and the standard benchmark Vietnamese Treebank \citep{nguyen-EtAl:2009:LAW-III} of about 10K sentences. 
 On these datasets, RDRPOSTagger\textsubscript{+TnT} achieves competitive results (93.63\% and 92.95\%) in comparison to MarMoT (94.13\% and 93.53\%). 
 
  In addition, on the Vietnamese Treebank, RDRPOSTagger  with the accuracy 92.59\% outperforms  the previously reported Maximum Entropy Model, Conditional Random Fields and Support Vector Machine-based approaches \citep{Tran09} where the highest obtained accuracy was 91.64\%.

  \subsection{Training time and tagging speed}
\label{ssec:tttg}

While most published works have not reported training times and tagging speeds, we present our single-threaded implementation results in Tables \ref{tab:acc1} and \ref{tab:acc2}.\footnote{To measure the tagging speed on a test fold, we perform the tagging process on the test fold 10 times and then take the average.} From there we can see that TnT is the fastest in terms of both training and tagging when compared to our RDRPOSTagger and MarMoT. 
Our  RDRPOSTagger and MarMoT  require similar training times, however, RDRPOSTagger is significantly faster than MarMoT in terms of tagging speed. 

It is interesting to note that in some languages, training our RDRPOSTagger  is faster for combined POS+MORPH tagging task than for POS tagging, as presented in experimental results for French (9 minutes vs 16 minutes) and German (22 minutes vs 28 minutes)   in Table \ref{tab:acc1}. Usually in machine learning-based approaches fewer number of tags leads to higher training speed. 
For example, on a 40,474-sentence subset of the German TIGER corpus \citep{Brants04}, SVMTool took about 899 minutes (about 15 hours) to train using 54 POS tags as compared to about  1,649 minutes (about 27 hours) using 681 POS+MORPH tags  \citep{mueller-schmid-schutze:2013:EMNLP}.

\begin{table}[ht]
\setlength{\tabcolsep}{0.25em}
\resizebox{7.5cm}{!}{
\begin{tabular}{lllllll}
\hline 
\bf Language & \bf \#sent  & \bf \#tags & SVMT & Morf & CRFS & RFT \\ 
\hline 
\hline
German & 40,474  & 681 & 1,649 & 286 & 1,295 & 5\\
\hline
Czech & 38,727 &  1,811 & 2,454 & 539 & 9,274 & 3\\
\hline
Spanish & 14,329  & 303 & 64 & 63 & 69 & 1\\
\hline
\hline
\end{tabular} 
}
\caption{The training time in minutes reported by \citet{mueller-schmid-schutze:2013:EMNLP} for POS+MORPH tagging on a machine of two Hexa-Core Intel Xeon X5680 CPUs with 3,33 GHz and 144 GB of memory. {\bf \#sent}: the number of sentences in training set. {\bf \#tag}: the number of POS+MORPH tags. SVMT: SVMTool, Morf:  Morfette, CRFS: CRFSuite, RFT: RFTagger.} 
\label{tab:trainingtime}
\end{table}

 In order to compare with other existing POS taggers in terms of the training time, we show in Table  \ref{tab:trainingtime} the time taken to train the SVMTool, CRFSuite,  Morfette and RFTagger  using a more powerful computer than ours. For instance, on the German TIGER corpus, RDRPOSTagger took an average of 22 minutes to train  a POS+MORPH tagging model while SVMTool and  CRFSuite  took   1,649 minutes (about 27 hours) and 1,295 minutes (about 22 hours) respectively, as shown in Table  \ref{tab:trainingtime}. Furthermore, RDRPOSTagger uses larger  datasets for Czech and Spanish and obtains  faster training process as compared to SVMTool, CRFSuite and  Morfette.

Regarding to tagging speed, as reported by \citet{Moore2014} using the same evaluation scheme on English  on a Linux workstation equipped with Intel Xeon X5550 2.67 GHz: the SVMTool, the UPenn bidirectional tagger \citep{Shen07}, the COMPOST tagger \citep{Spoustova2009}, \citet{Moore2014}'s approach, the accurate version of the Stanford tagger  \citep{Toutanova2003} and the fast and less accurate version of the Stanford tagger   gained tagging speed of 7700, 270, 2600, 51K, 5900 and 80K tokens per second, respectively.  In our experiment, RDRPOSTagger obtains a faster tagging speed of 279K tokens per second on a weaker computer. To the best of our knowledge, we conclude that RDRPOSTagger is fast both in terms of training and tagging in comparison to other approaches.

\section{Related work}
\label{sec:relatedwork}

From early POS tagging approaches the rule-based Brill's tagger \citep{Brill95} is the most well-known. 
The key idea of the Brill's method is to compare a manually annotated gold standard corpus  with an initialized corpus which is generated by executing an initial tagger on the corresponding unannotated corpus. Based on the predefined rule templates, the method then automatically produces a list of concrete rules to correct wrongly assigned POS tags. 
For example, the template \textit{``transfer tag of current word from \textbf{A} to \textbf{B} if the next word is \textbf{W}''} can produce concrete rules  such as  \textit{``transfer tag of current word from \textbf{JJ} to \textbf{NN} if the next word is \textbf{of}''} or \textit{``transfer tag of current word from \textbf{VBD} to \textbf{VBN} if the next word is \textbf{by}.''}

At each training iteration, the Brill's tagger generates a set of all possible rules and chooses the ones that help to correct the incorrectly tagged words in the whole corpus. Thus, the Brill's training process  takes a significant amount of time. 
To prevent that,  \citet{Hepple00} presented an approach with two assumptions for disabling interactions between rules to reduce the training time while sacrificing a small amount of accuracy. \citet{Ngai01} proposed another method to  reduce the training time by recalculating the scores of  rules while obtaining similar accuracy result. 

The \textit{main difference} between our approach and the Brill's method is that we construct transformation rules in the form of a SCRDR tree where a new transformation rule is produced only based on  a subset of  tagging errors.  So our approach is faster  in term of  training speed. In the conference publication version of our approach  \citep{nguyen2011ripple}, we reported an improvement up to 33 times in training speed against the Brill's method. 
In addition, the Brill's method enables each subsequent rule to change the outputs  of all preceding rules, thus a word can be tagged multiple times in the tagging process, each time by a different rule. This is different from our approach where each word is tagged only once. Consequently, our approach also achieves a faster tagging speed.

In addition to our research, there  is only one work that applies Ripple Down Rules method for POS tagging  proposed by \citet{Xu2010}. Though Xu
and Hoffmann's method obtained a very competitive accuracy, it is a hand-crafted approach taking about 60 hours to manually build a SCRDR tree model for English POS tagging. 

Turning to statistical and machine learning methods for POS tagging, these methods can be listed as various Hidden Markov model-based methods \citep{Brants00,Fruzangohar2013,Thede1999}, maximum entropy-based methods \citep{CHRUPALA08,Ratnaparkhi1996,Tagger2000,Toutanova2003,Tsuruoka05}, perceptron algorithm-based approaches \citep{Collins02,Shen07,Spoustova2009}, neural network-based approaches \citep{Carneiro:2015:MPT:2779640.2779998,Collobert:2011:NLP:1953048.2078186,labeau-loser-allauzen:2015:EMNLP,ma-zhang-zhu:2014:P14-1,icml2014c2_santos14,Schmid1994,zheng-chen-xu:2013:EMNLP}, Conditional Random Fields \citep{Lafferty01,Lavergne:2010:PVL:1858681.1858733,li-EtAl:2015:ACL-IJCNLP3,mueller-schmid-schutze:2013:EMNLP,muller-EtAl:2015:EMNLP}, Support Vector Machines \citep{Gimenez04,kim-snyder-sarikaya:2015:EMNLP,TACL183,song-EtAl:2012:ACL2012}  and other  approaches including decision trees \citep{Schmid1994a,Schmid:2008} and hybrid methods \citep{Forsati2014,Lee:2002:SUS:636735.636739}. Overview about the POS tagging task can be found in \citep{Gungor2010,Horsmann2015}.

\section{Conclusion and future work}
\label{sec:Conclusion}

In this paper, we propose a new error-driven method to automatically construct a Single Classification Ripple Down Rules tree of transformation rules for POS and morphological tagging. Our method allows the interaction between rules where a rule only changes the results of a limited number of other rules. Experimental evaluations for POS tagging and the combined POS and morphological tagging on 13  languages show that our method obtains very promising accuracy results. In addition, we successfully achieve  fast training and tagging processes for all experimental languages. 
This could help to significantly reduce time and effort for the machine learning tasks on big data, employing POS and morphological information as learning features.  

An important point is that our approach is  suitable to involve domain experts to add new exception rules given  concrete cases that are misclassified by the tree model. This is especially important for under-resourced languages where obtaining a large annotated corpus is difficult.  
In future work, we plan to build tagging models for  other languages such as Russian, Arabic, Latin, Hungarian, Chinese and so forth.

\section*{Bibliographic note}
  
This paper extends the work published in our conference publications \citep{nguyen2011ripple,Nguyen2014a}. We make minor revisions to our published approach to yield improved accuracy results on English and Vietnamese, and we conduct new extensive empirical study on 11 other languages.

\section*{Acknowledgments}
  This research work is partially supported by the research project ``VNU-SMM: An automated system for monitoring online social media to assist management and support decision making in business, politics, education, and social areas''  from Vietnam National University, Hanoi. The first author is supported by an International Postgraduate Research Scholarship and a NICTA NRPA Top-Up Scholarship. 
The authors would like to thank the three anonymous reviewers, the associate editor Prof. Fabrizio Sebastiani and  Dr. Kairit Sirts at the Macquarie University, Australia for helpful comments and suggestions.

{\small
\bibliographystyle{abbrvnat}

\bibliography{POSTagRefer}
}

\end{document}